\documentclass[lettersize,journal]{IEEEtran}
\usepackage{amsmath,amsfonts}
\usepackage{algorithmic}
\usepackage{algorithm}
\usepackage{array}
\usepackage[caption=false,font=normalsize,labelfont=sf,textfont=sf]{subfig}
\usepackage{textcomp}
\usepackage{stfloats}
\usepackage{url}
\usepackage{verbatim}
\usepackage{graphicx}
\usepackage{cite}
\usepackage{amssymb} 
\usepackage{booktabs}
\usepackage{multirow}
\usepackage{pifont}
\usepackage[dvipsnames,table,xcdraw]{xcolor}
\definecolor{mygray}{gray}{.92}
\usepackage[colorlinks,
            linkcolor=blue,      
            anchorcolor=blue,  
            citecolor=blue,        
            ]{hyperref}

\begin{document}

\title{${D}^{3}${ETOR}: ${D}$ebate-Enhanced Pseudo Labeling and Frequency-Aware Progressive ${D}$ebiasing for Weakly-Supervised Camouflaged Object ${D}$etection with Scribble Annotations}

\author{Jiawei~Ge, Jiuxin~Cao, Xinyi~Li, Xuelin~Zhu, Chang~Liu, Bo~Liu, Chen~Feng, \\ 
and Ioannis~Patras,~\IEEEmembership{Senior Member,~IEEE}
\IEEEcompsocitemizethanks{
\IEEEcompsocthanksitem{This work is supported by National Natural Science Foundation of China under Grants No.62472092, No. 62172089, No.62106045. Natural Science Foundation of Jiangsu Province under Grants No. BK20241751. Jiangsu Provincial Key Laboratory of Computer Networking Technology. Jiangsu Provincial Key Laboratory of Network and Information Security under Grants No. BM2003201, and Key Laboratory of Computer Network and Information Integration of Ministry of Education of China under Grants No. 93K-9, Nanjing Purple Mountain Laboratories, Fintech and Big Data Laboratory of Southeast University. We thank the Big Data Computing Center of Southeast University for providing the facility support on the numerical calculations. (\textbf{Corresponding author: Jiuxin Cao})}
\IEEEcompsocthanksitem{J.~Ge, J.~Cao, X.~Li, and C.~Liu are with School of Cyber Science and Engineering, Southeast University, Nanjing 211189, China.
(e-mail: jiawei\_ge@seu.edu.cn, jx.cao@seu.edu.cn, lLydiaxy@outlook.com, liuchang22@seu.edu.cn)} 
\IEEEcompsocthanksitem{X.~Zhu is with Department of Aeronautical and Aviation Engineering, The Hong Kong Polytechnic University, Hong Kong, China (e-mail: xuelin.zhu@polyu.edu.hk} 
\IEEEcompsocthanksitem{B.~Liu is with School of Computer Science and Engineering, Southeast University, Nanjing 211189, China. 
(e-mail: bliu@seu.edu.cn)}  
\IEEEcompsocthanksitem{C.~Feng is with School of Electronics, Electrical Engineering and Computer Science, Queen's University Belfast, Belfast, U.K.
(e-mail: c.feng@qub.ac.uk)}  
\IEEEcompsocthanksitem{I.~Patras is with School of Electronic Engineering and Computer
Science, Queen Mary University of London, London E1 4NS, U.K. 
(e-mail: i.patras@qmul.ac.uk)}  
}
}

\maketitle

\begin{abstract}
Weakly-Supervised Camouflaged Object Detection (WSCOD) aims to locate and segment objects that are visually concealed within their surrounding scenes, relying solely on sparse supervision such as scribble annotations. Despite recent progress, existing WSCOD methods still lag far behind fully supervised ones due to two major limitations: (1) the pseudo masks generated by general-purpose segmentation models (e.g., SAM) and filtered via rules are often unreliable, as these models lack the task-specific semantic understanding required for effective pseudo labeling in COD; and (2) the neglect of inherent annotation bias in scribbles, which hinders the model from capturing the global structure of camouflaged objects. To overcome these challenges, we propose ${D}^{3}$ETOR, a two-stage WSCOD framework consisting of Debate-Enhanced Pseudo Labeling and Frequency-Aware Progressive Debiasing. In the first stage, we introduce an adaptive entropy-driven point sampling method and a multi-agent debate mechanism to enhance the capability of SAM  for COD, improving the interpretability and precision of pseudo masks. In the second stage, we design FADeNet, which progressively fuses multi-level frequency-aware features to balance global semantic understanding with local detail modeling, while dynamically reweighting supervision strength across regions to alleviate scribble bias. By jointly exploiting the supervision signals from both the pseudo masks and scribble semantics, ${D}^{3}$ETOR significantly narrows the gap between weakly and fully supervised COD, achieving state-of-the-art performance on multiple benchmarks.
\end{abstract}


\section{Introduction}

\begin{figure}[]
    \centering
    \includegraphics[width=\linewidth]{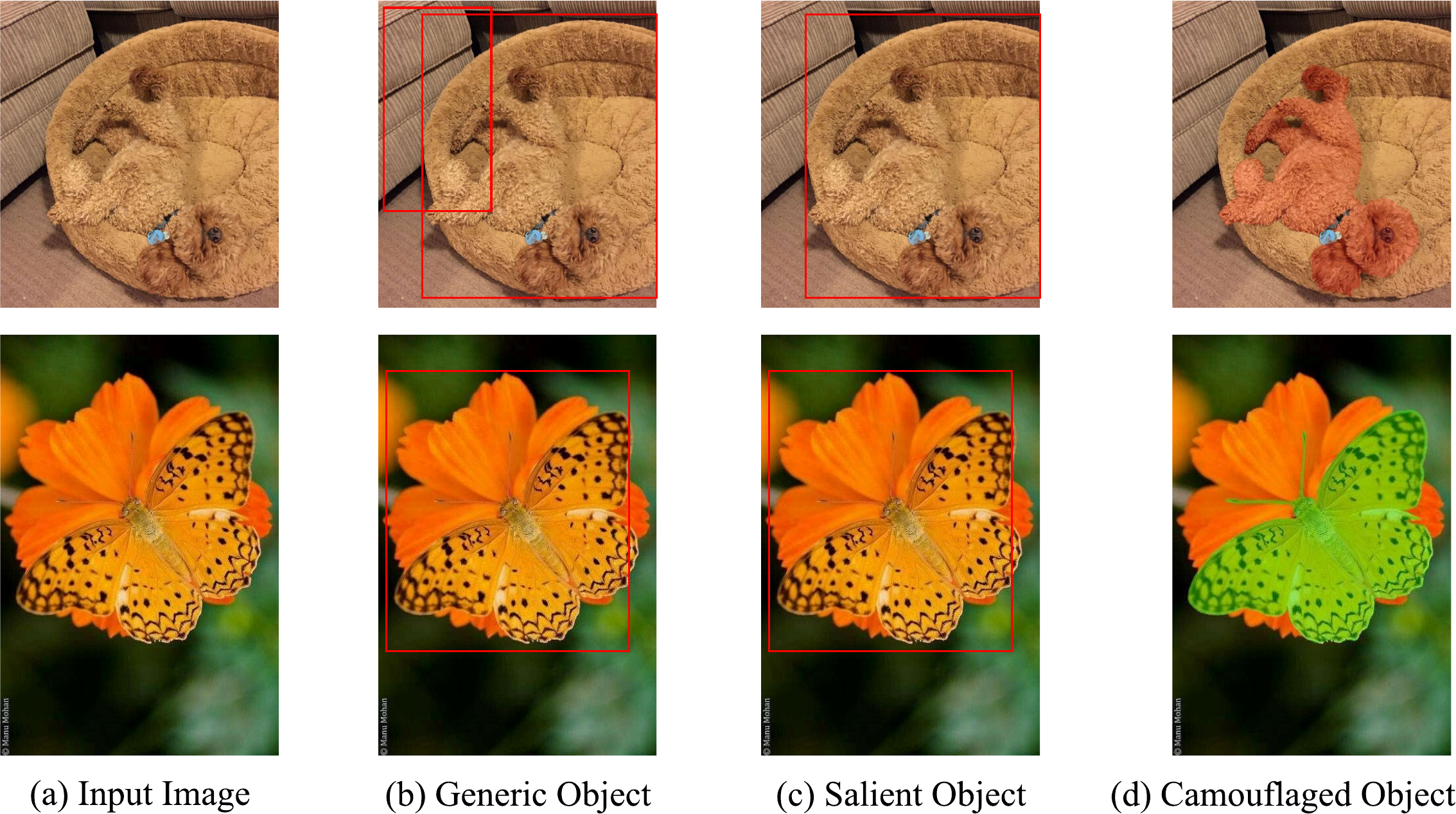}
    \caption{Visualization of task objectives across different detection paradigms. (a) Input raw image. (b) Generic object detection aims to locate and classify all identifiable objects (segmented with distinct colors). (c) Salient object detection focuses on identifying objects that naturally attract human attention (e.g., the stick highlighted in green). (d) Camouflaged object detection identifies visually inconspicuous objects that blend into the background, such as the small snake highlighted in red.}
    \label{diff}
\end{figure}

\IEEEPARstart{C}{amouflaged} Object Detection (COD) is a challenging task that involves identifying and segmenting objects that are well-concealed within their environments \cite{fan2020camouflaged}. In contrast to salient object detection \cite{wang2021salient} or generic object detection \cite{liu2020deep}, COD must handle scenarios where an object’s appearance, shape, and color are almost indistinguishable from the background (as shown in Figure \ref{diff} (d)). These objects range from animals exhibiting protective coloration or small, occluded forms to man-made objects engineered for concealment. The high similarity to the surrounding background, along with indistinct boundaries and deceptive textures, poses significant challenges for detection. Despite these challenges, COD still attracts growing interest in the computer vision community and shows promise in applications such as species discovery \cite{pang2022zoom,feng2025prosac}, agriculture monitoring \cite{chudzik2020mobile,feng2026deconstructing, feng2023maskcon}, medical image segmentation \cite{fan2020pranet,fan2020inf,wu2021jcs}, and wildlife tracking \cite{fan2021concealed,zhu2026autoit,zhu2025mambaml,ge2025gen4track, ge2024consistencies, ge2025beyond, wang2025r1trackdirectapplicationmllms,ge2025fsd,ge2024gal}.

While existing COD methods have achieved impressive performance, they rely heavily on dense pixel-wise annotations, which can take up to 60 minutes per image \cite{fan2020camouflaged}. In contrast, weakly-supervised labels offer a much more efficient and scalable alternative. For instance, scribble-based annotations require only around 10 seconds per image \cite{he2023weakly}. As a recently emerging task, Weakly-Supervised Camouflaged Object Detection (WSCOD) has seen only limited exploration \cite{he2023weakly,hec2023weakly,chen2024sam}. Early efforts such as CRNet \cite{he2023weakly} focus on expanding the scribbles to wider camouflaged regions to capture structural information and semantic relationships. More recent methods, WS-SAM \cite{hec2023weakly} and SAM-COD \cite{chen2024sam}, push the boundary of WSCOD by adopting a two-stage learning framework: they first generate high-quality pseudo masks from scribbles using the Segment Anything Model (SAM) \cite{kirillov2023segment}, and then train COD models on these pseudo masks. Despite these advances, current WSCOD approaches still lag far behind fully supervised COD models in terms of detection accuracy and segmentation quality. We observe that such performance gaps primarily result from the suboptimal exploitation of weak supervision signals, particularly in two key aspects: \textbf{(1) the limited quality and quantity of supervision signals generated by SAM}, and \textbf{(2) the harmful annotation bias inherent in scribbles introduced by annotators}.

\begin{figure}[]
    \centering
    \includegraphics[width=\linewidth]{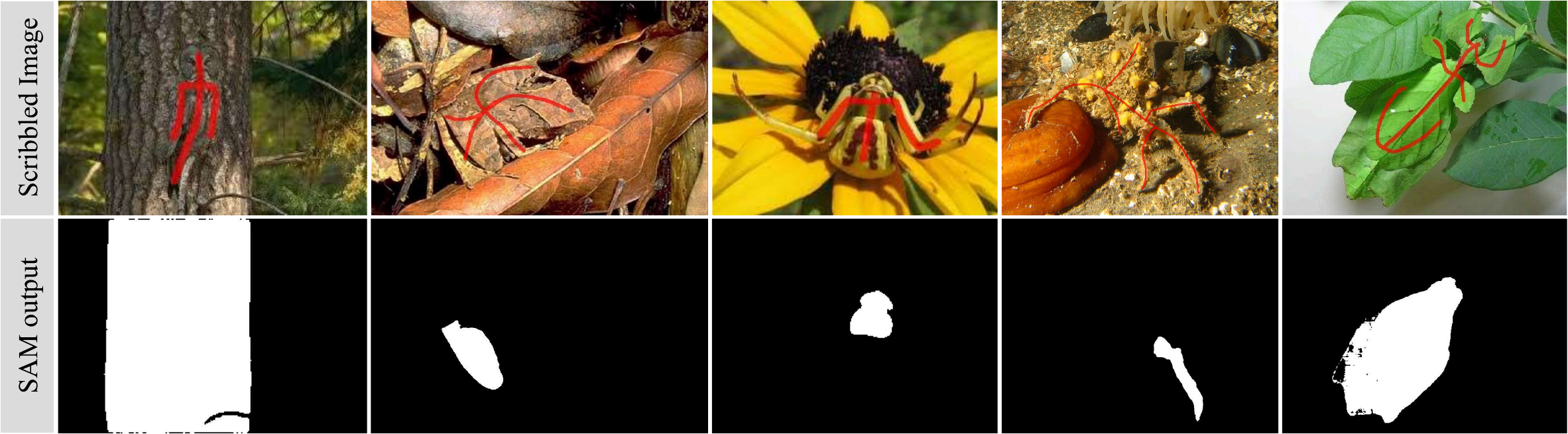}
    \caption{As a general-purpose segmentation model, SAM struggles to meet the specific demands of the COD task. Consequently, its limited semantic discriminative capability often leads to suboptimal segmentation performance on camouflaged objects.}
    \label{issue}
\end{figure}

To begin with, effectively leveraging the Segment Anything Model (SAM) \cite{kirillov2023segment} for the WSCOD task is non-trivial. Although SAM can directly generate pseudo masks for generic objects, it is prone to producing unwanted semantic responses, such as non-camouflaged objects or fragmented local regions (as shown in Figure \ref{issue}). This limitation is largely due to SAM’s restricted semantic understanding and its inability to comprehend the specific demands of camouflaged object detection. To alleviate this issue, prior works like WS-SAM \cite{hec2023weakly} and SAM-COD \cite{chen2024sam} propose heuristic selection mechanisms with predefined thresholds to filter unreliable pseudo masks. However, such strategies still \textbf{fail to equip SAM with the semantic reasoning capabilities necessary for COD. Ultimately, the quality of the pseudo masks remains constrained by the inherent limitations of SAM}. Inevitably, many \textbf{informative hard examples may be prematurely discarded due to low confidence scores in the rule-based selection}. As a result, the current exploitation of weak supervision in WSCOD is insufficient, both in quality and quantity, to support the training of robust and accurate models.

\begin{figure}[]
    \centering
    \includegraphics[width=0.8\linewidth]{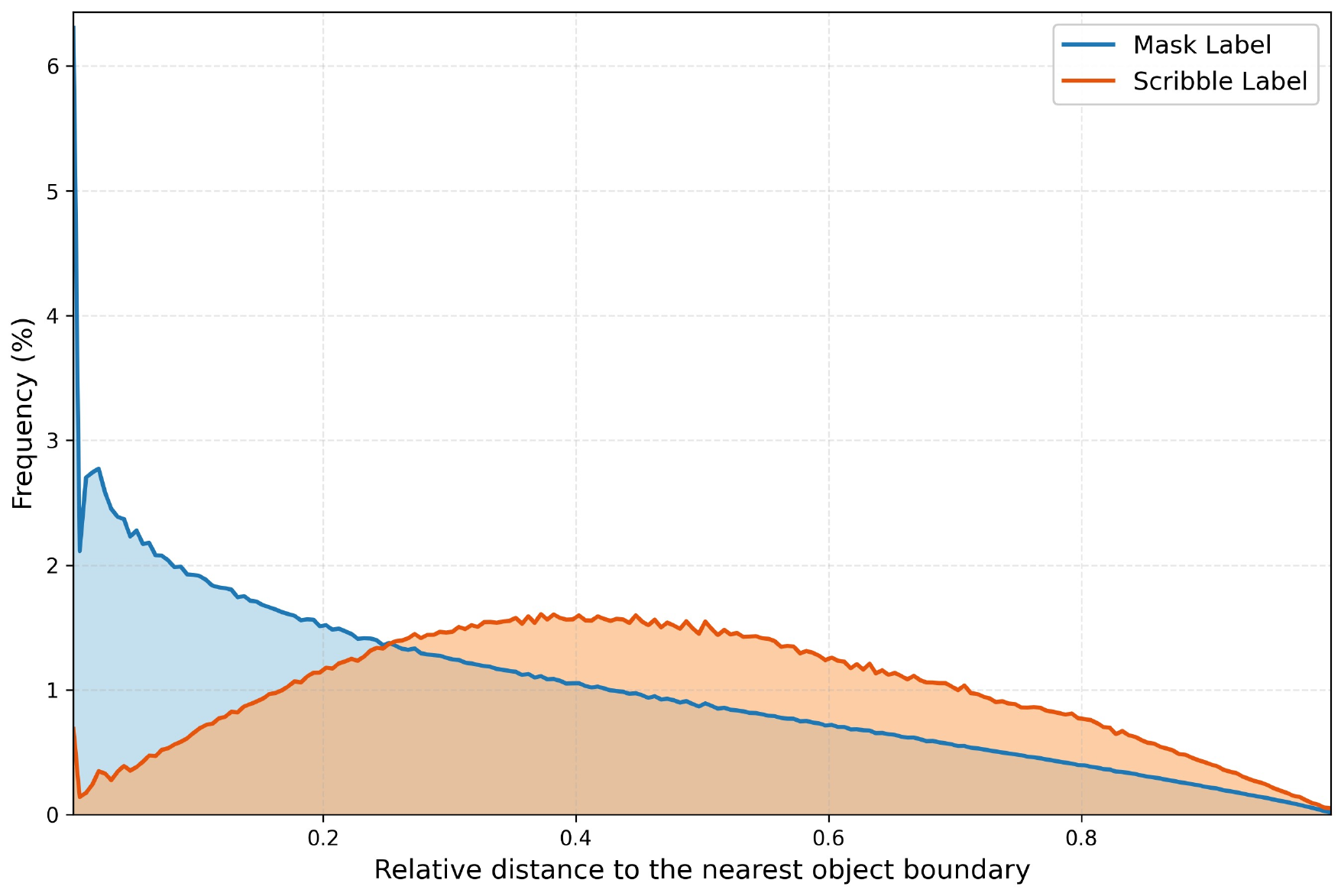}
    \caption{The relative distances from ground truth mask and scribble pixels to their respective nearest object boundaries reveal a notable discrepancy in spatial distribution. It is evident that scribble annotations are biased toward central object regions, whereas mask labels exhibit a more uniform coverage across entire object areas.}
    \label{dist}
\end{figure}

Furthermore, \textbf{these methods tend to overlook the rich semantic information (e.g., structure and relation) embedded in scribble annotations once they have pseudo masks, leaving such valuable information underexplored.} Although CRNet \cite{he2023weakly} tried to exploit the rich semantics contained in scribble annotations, it failed to recognize and tackle the harmful bias inherent in scribble annotations. As illustrated in Figure \ref{dist}, our statistical analysis reveals a notable discrepancy between the spatial distribution of scribble annotations and ground truth masks. This bias arises from annotators' preference for placing scribbles near the central regions of objects, leading to an imbalanced coverage of camouflaged objects. Consequently, \textbf{models trained on such biased supervision signals tend to perform well in frequently annotated regions but poorly in less emphasized areas}. Such bias hidden in the supervision signal limits the generalization ability of the WSCOD model and constrains the network’s discriminative capacity, ultimately hindering the accurate detection of camouflaged objects.

\begin{figure}[]
    \centering
    \includegraphics[width=\linewidth]{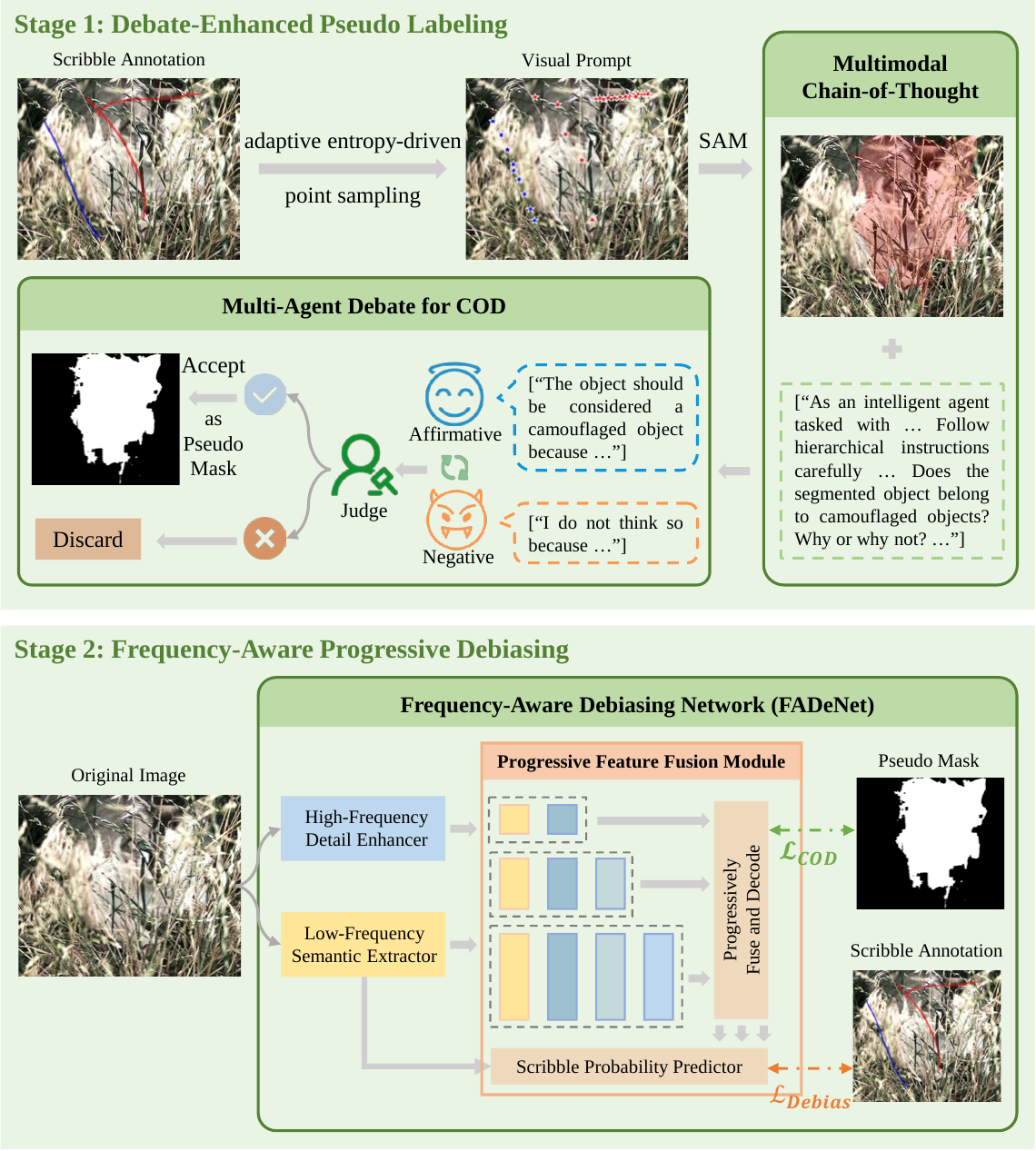}
    \caption{An overview of the proposed ${D}^{3}${ETOR} framework for weakly-supervised camouflaged object detection, which consists of two stages: debate-enhanced pseudo labeling and frequency-aware progressive debiasing.}
    \label{overview}
\end{figure}

To address the aforementioned challenges, we propose ${D}^{3}${ETOR}, a two-stage framework for weakly-supervised camouflaged object detection with scribble annotations. As illustrated in Figure \ref{overview}, $\mathrm{D}^{3}$ETOR first generates pseudo labels using SAM with a multi-agent debate strategy and then trains our Frequency-Aware Debiasing Network (FADeNet) to accurately detect camouflaged objects. Specifically, we design an adaptive entropy-driven point sampling method to generate visual prompts from scribble annotations, enabling SAM to efficiently produce pseudo masks of potential camouflaged objects. To prevent pseudo masks from containing noise (e.g., irrelevant objects or incomplete segmentation) or from discarding camouflaged objects when confidence is low, we draw insight from the fundamental characteristic of human problem solving (i.e., debate) and further introduce a multi-agent debate strategy based on the multimodal chain-of-thought to encourage divergent reasoning for pseudo label selection. This strategy improves the interpretability and precision of pseudo labeling, hence ensuring both the quality and quantity of pseudo masks used for training of downstream COD models.

In the second stage, we take inspiration from frequency-based methods \cite{sun2024frequency,lin2023frequency} and devise FADeNet, a Frequency-Aware Debiasing Network for COD, to progressively leverage supervision signals from both pseudo masks and scribble annotations. By rethinking COD from a frequency perspective, FADeNet extracts representative signals through a low-frequency semantic extractor and a high-frequency detail enhancer, thereby balancing global semantic understanding with local detail modeling. Furthermore, a progressive feature fusion module integrates multi-scale frequency-aware features with the window-based cross-attention mechanism, enabling a deeper comprehension of multi-level visual representations and significantly enhancing the accuracy of camouflaged object detection. To better utilize the rich semantics (e.g., structural and relational cues) in scribble annotations while mitigating their inherent harmful bias, we introduce an auxiliary debiasing task during training. By dynamically adjusting supervision strength across regions according to the scribble probability, this task guides the model’s attention toward areas that are difficult to annotate or are easily overlooked, improving overall structure perception and enhancing generalization ability under the weakly-supervised learning setting.

In summary, the contributions of our ${D}^{3}${ETOR} can be detailed as follows:

\begin{enumerate}
    \item We present ${D}^{3}${ETOR}, a two-stage framework for WSCOD with scribble annotations that strengthens the pseudo labeling pipeline through multi-agent debate and mitigates annotation bias in scribbles via a frequency-aware debiasing network. By fully exploiting weak supervision signals from scribble annotations, ${D}^{3}${ETOR} effectively narrows the gap between fully supervised and weakly-supervised camouflaged object detection.

    \item To better utilize the rich semantics in scribble annotations under weak supervision, we propose a debate-enhanced pseudo labeling strategy that strengthens SAM’s capacity for COD, efficiently producing higher-quality and abundant pseudo masks for subsequent training.

    \item We further design FADeNet, a frequency-aware debiasing network that progressively fuses multi-scale and multi-level frequency representations through window-based cross-attention, while dynamically adjusting supervision strength to better capture overall structural information.

    \item Extensive experiments on multiple COD datasets demonstrate that our framework consistently achieves SOTA performance under weak supervision, validating both its effectiveness and generalization ability.
\end{enumerate}

\section{Related Work}
\subsection{Camouflaged Object Detection}
Camouflaged object detection (COD), which aims to identify and segment objects that blend seamlessly with their surroundings, has attracted growing attention in the computer vision community. SINet \cite{fan2020camouflaged} pioneered COD by introducing a search-and-identification network inspired by the animal predation mechanism, while SINetv2 \cite{fan2021concealed} improved localization ability through texture enhancement modules and cascade attention. FEDER \cite{he2023camouflaged} decomposed features into frequency bands and reconstructed edges using an ordinary differential equation-inspired module, striving to address target-background similarity. Transformer-based methods further advanced visual representation learning in COD. FSPNet \cite{huang2023feature} proposed a feature shrinkage pyramid network with non-local token to enhance feature modeling and aggregation. MSCAF-Net \cite{liu2023mscaf} integrated multi-scale context and cross-scale fusion for progressive camouflaged object detection, whereas CamoFormer \cite{yin2024camoformer} employed mask-separable attention and top-down decoding to enhance segmentation quality. More recently, CamoDiffusion \cite{chen2024camodiffusion} adopted a diffusion-based denoising process to iteratively refine predictions, mitigating the overconfident problem in mis-segmentation results. MCRNet \cite{zhang2025mamba} abstracted type-level capsules from pixel-level features to reduce computational cost while preserving part-whole relationships.

In spite of these progress, most methods rely on large-scale datasets with pixel-level annotations. The inherently ambiguous boundaries of camouflaged objects make such pixel-wise labeling both time-consuming and labor-intensive. To address this, He et al. \cite{he2023weakly} introduced the S-COD dataset, which provides scribble annotations as a weak supervision signal. WS-SAM \cite{hec2023weakly} generated pseudo masks from sparse cues and applied multi-scale feature grouping with entropy-based weighting to enhance reliability. SAM-COD \cite{chen2024sam} proposed a response filter to remove extreme responses from SAM via computing the ratio of the mask to the image size, along with a semantic matcher to select masks that balance the segmentation detail and semantic accuracy. However, these approaches still solely rely on manually designed strategies to filter pseudo labels generated by SAM, which is a general-purpose model and cannot capture the task-specific requirements of COD. As a result, their performance remains substantially inferior to that of fully supervised COD models. To bridge this gap, we propose ${D}^{3}${ETOR} to enhance the existing pseudo labeling pipeline through multi-agent debating, hence empowering SAM with task-aware reasoning ability for COD. In addition, we introduce a frequency-aware debiasing network in the later stage of training to mitigate annotation bias in scribbles, further improving segmentation accuracy and completeness.

\subsection{Chain-of-Thought Prompting}
The idea of prompt learning was first introduced in natural language processing (NLP), where handcrafted prompts were designed to motivate task-specific behaviors from large language models \cite{brown2020language}. Following its success, the paradigm was soon extended to computer vision, with applications in recognition, segmentation, and vision-language modeling \cite{jia2022visual,khattak2023maple}. A related advancement, Chain-of-Thought (CoT) prompting \cite{wei2022chain}, showed that decomposing complex reasoning into intermediate steps can significantly enhance model performance. This principle has also been adopted in visual domains, such as segmentation and structured reasoning \cite{lu2022learn,li2024cpseg}. Early research in this area primarily focused on prompt design and decoding strategies, but recent work has shifted toward mechanisms for iterative self-improvement. For example, Self-Refine \cite{madaan2023self} and Tree of Thoughts \cite{yao2023tree} allow models to generate candidate solutions and subsequently refine them through self-evaluation. While effective, these approaches remain constrained to the single-agent paradigm, which is inherently vulnerable to error accumulation and lacks robust self-correction.

To address these limitations, latest studies have proposed multi-agent debate frameworks, where multiple agents generate, critique, and refine outputs through interactive reasoning. This setup has demonstrated improved error correction and more consistent outcomes compared to single-agent CoT \cite{xiong2023diving,du2023improving}. Motivated by these advances, we pioneer the integration of the multi-agent debate paradigm into weakly-supervised camouflaged object detection (WSCOD). Our method leverages reasoning and collaboration among agents to boost SAM’s capacity for COD, generating pseudo masks that are both more accurate and reliable for training purposes.

\section{Proposed Method}

\begin{figure*}[]
    \centering
    \includegraphics[width=\linewidth]{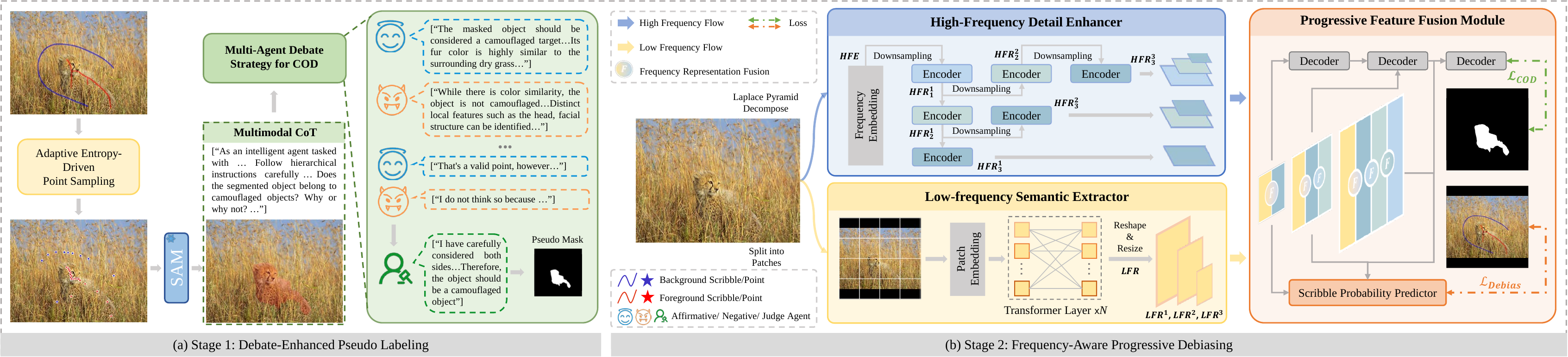}
    \caption{Framework of our proposed ${D}^{3}${ETOR} for weakly-supervised camouflaged object detection (WSCOD) with scribble annotations. In (a), candidate masks are first generated using visual-prompted SAM and then filtered through a multi-agent debate mechanism. Afterwards, images are decomposed into low-frequency and high-frequency components in (b), balancing global semantics and local details. These features are progressively fused via window-based cross-attention mechanism to refine segmentation results, while supervision strength across regions is dynamically adjusted across regions to mitigate harmful bias in scribble annotations.}
    \label{framework}
\end{figure*}

Training a model for weakly-supervised camouflaged object detection (WSCOD) can be challenging, as sparse scribble annotations alone often fail to provide sufficient supervision compared to dense mask labels. Although pseudo masks can be generated to mitigate this limitation, the original scribbles still contain valuable semantic cues that should not be ignored. However, directly incorporating them into training may introduce harmful bias, since annotators tend to place scribbles near central regions of objects, leading to imbalanced object coverage. To address these issues, we introduce ${D}^{3}${ETOR}, a two-stage framework for WSCOD. As illustrated in Figure \ref{framework}, the first stage (Sec. \ref{s1}) employs debate-enhanced pseudo labeling, where the multi-agent reasoning converts sparse scribbles into high-quality pseudo masks. In the second stage (Sec. \ref{s2}), we progressively extract and fuse frequency-aware representations with the window-based cross-attention, while simultaneously balancing annotation bias during training. In the following, we elaborate on the design and implementation of each stage in detail.

\subsection{Debate-Enhanced Pseudo Labeling}\label{s1}
In the first stage, we improve the SAM-based pseudo labeling pipeline to generate abundant high-quality pseudo masks from the semantic cues embedded in scribble annotations, strengthening the interpretability and precision of pseudo masks for subsequent learning. As shown in Figure \ref{framework} (a), this stage mainly consists of two components: (1) an adaptive entropy-driven point sampling method that produces effective visual prompts for SAM, and (2) a multi-agent debate strategy equipped with multimodal Chain-of-Thought (CoT) reasoning to filter candidate masks.

\subsubsection{\textbf{Adaptive Entropy-Driven Point Sampling}}
As a powerful vision foundation model for generic object segmentation, SAM \cite{kirillov2023segment} still faces challenges in detecting camouflaged objects \cite{chen2023sam} due to the high visual similarity between foreground and background. To achieve accurate pseudo labeling for COD, SAM requires visual prompts that explicitly indicate target objects. However, it only supports prompts such as points, masks, and bounding boxes, while scribble-type inputs are not directly compatible. To address this issue, we propose an Adaptive Entropy-Driven Point Sampling strategy that converts sparse scribbles into informative visual prompts for SAM, thereby fully leveraging its capacity for COD.  

The process begins with estimating the local entropy of scribble regions, which quantifies the structural uncertainty of each pixel. Specifically, for a pixel $(x,y)$ within foreground or background scribbles $S$, the local entropy is defined as
\begin{equation}
H(x, y) = - \sum_{i}^{} p_i \log p_i, 
\quad 
p_i = \frac{h_i}{\sum_j h_j}, \; p_i > 0,
\end{equation}
where $h_i$ denotes the histogram count of intensity value $i$ within a local neighborhood. Pixels with higher entropy correspond to more informative yet ambiguous regions that are crucial for segmentation. Candidate points are then selected as
\begin{equation}
\mathcal{C} = \{p \in \mathcal{S} \mid H(p) \geq \tau \cdot H_{\max} \},
\end{equation}
where $\tau$ is a normalized entropy threshold.  

To prevent over-clustering in high-entropy areas, we apply spatial filtering by enforcing a minimum distance constraint:
\begin{equation}
\|p_i - p_j\| > d_{\min}, \quad \forall p_j \in S_\text{selected}.
\end{equation}
This ensures that selected points remain well-separated while preserving local structural information.  

Finally, to guarantee uniform coverage across scribble annotations, we incorporate {Farthest Point Sampling (FPS) \cite{eldar1997farthest}}, which iteratively selects $N$ points that maximizes its minimum distance to the already chosen set:
\begin{equation}
p_k = \arg\max_{i} \min_{j < k} \|p_i - p_j\|, \quad k=2,\dots,N.
\end{equation}

In this way, we obtain visual prompts that are both informative and spatially balanced, enabling SAM to generate more reliable pseudo masks for WSCOD.

\begin{figure}[]
    \centering
    \includegraphics[width=\linewidth]{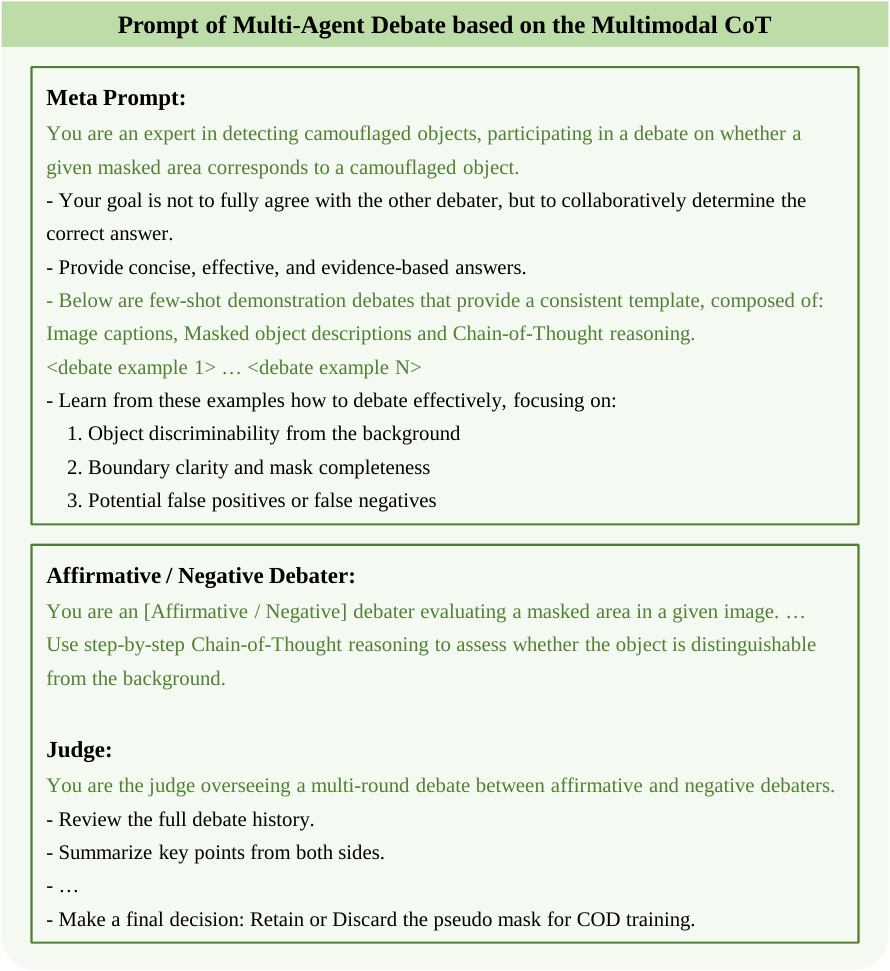}
    \caption{The prompt examples in our Multi-Agent Debate strategy.}
    \label{prompt}
\end{figure}

\subsubsection{\textbf{Multi-Agent Debate based on the Multimodal Chain-of-Thought}}
Since SAM is not fine-tuned for COD due to the lack of sufficient supervision signals in the weakly-supervised learning setting, the pseudo masks it generates often contain noise, and many informative hard examples may be discarded because of low confidence scores. To address this limitation, we introduce a Multi-Agent Debate (MAD) strategy based on the multimodal Chain-of-Thought to encourage divergent thinking for COD. In this strategy, agents powered by the Multimodal Large Language Model (MLLM) iteratively generate supportive and opposing reasoning chains for each image–mask pair, while a judge agent evaluates these chains-of-thought to decide whether the mask should be retained as a pseudo label for COD. Through this process, our approach improves the existing pipeline in WSCOD for pseudo label selection, in both interpretability and precision.

As shown in Figure , our MAD strategy consists of three key components, described as follows:

\noindent
\textbf{Meta Prompt.} We employ meta prompts to introduce the COD task, iteration limits, and other specific requirements. To ensure that the reasoning chains capture camouflage-specific characteristics \cite{he2023weakly}, we design representative exemplars \cite{wei2022chain} as paradigm cases. These exemplars are refined with human priors and expanded into few-shot demonstration chains, providing a consistent template composed of image captions, masked object descriptions, and Chain-of-Thought (CoT) reasoning. The paradigm chains specify how pro and con viewpoints should evaluate attributes such as discriminability and boundary clarity, efficiently guiding the reasoning process of MLLMs.

\noindent
\textbf{Debaters.} Two types of debaters participate in each debate: affirmative and negative. During each iteration, debaters take turns speaking in a fixed order, formulating their arguments based on the accumulated debate history. This iterative exchange encourages divergent reasoning and thorough exploration of potential pseudo label correctness.

\noindent
\textbf{Judge.} A judge agent is designed to observe the entire debate process. The judge evaluates the complete debate history and determines the final decision on whether a pseudo mask should be retained for COD training.

\subsection{Frequency-Aware Progressive Debiasing}\label{s2}
Due to the high similarity between camouflaged objects and their backgrounds, accurately delineating object boundaries remains a challenging task for WSCOD models, even when pseudo masks generated by SAM are available from the first stage. Moreover, the inherent bias in scribble annotations poses additional challenges when they are used as supervision signals, impairing model optimization during training. Fortunately, previous studies \cite{sun2024frequency,lin2023frequency} have shown that these boundaries can be distinguished more effectively when analyzed in the frequency domain. Motivated by these insights, we rethink COD from the perspective of frequency components and propose FADeNet, a Frequency-Aware Debiasing Network for COD, to progressively leverage supervision signals from both pseudo masks and scribble annotations.

The architecture of FADeNet is illustrated in Figure \ref{framework} (b). Given an RGB image, the Low-Frequency Semantic Extractor captures low-frequency features via self-attention to model the global context of camouflaged scenes. High-frequency features are obtained through Laplacian pyramid decomposition, followed by multi-level and multi-scale frequency encoding within the High-Frequency Detail Enhancer, preserving detailed information across frequency components. Subsequently, our method progressively fuses these frequency representations with window-based cross-attention to reduce discrepancies across frequencies and hierarchical levels, generating features that integrate both contextual and fine-grained information. Concurrently, a scribble probability predictor highlights the preferences of scribble annotations, allowing us to balance label bias during training. Finally, the CNN-based decoder fuses these hierarchical feature representations in a bottom-up manner to produce the final prediction.

\subsubsection{\textbf{Frequency-Aware Encoding}}

In camouflaged object detection, the primary challenge lies in suppressing interference from visually confusing backgrounds while accurately delineating object boundaries. Although high-frequency components help reduce visual distraction by emphasizing edge details, they inherently lack sufficient semantics, leading to an incomplete contextual understanding. To overcome this limitation, we propose a frequency-aware encoding method that hierarchically leverages frequency representations across multiple levels, enabling complementary feature learning from both low- and high-frequency domains.

\noindent
\textbf{Low-Frequency Semantic Extractor (LFSE).} 
We adopt a Vision Transformer (ViT) as the low-frequency encoder to extract global semantic representations from the image, as previous studies \cite{park2022vision,paul2022vision} have demonstrated that the multi-head self-attention mechanism in transformer encoders naturally captures low-frequency components, effectively functioning as a low-pass filter. Specifically, the input image $I \in \mathbb{R}^{H \times W \times 3}$ is first divided into non-overlapping patches of size $16 \times 16$. Each patch is then flattened and linearly projected into a $D$-dimensional embedding, forming a token sequence that represents local visual information. After processing through $n$ Transformer layers, we obtain the low-frequency representation $LFR \in \mathbb{R}^{\frac{H}{16} \times \frac{W}{16} \times D}$, which effectively captures the global semantics of the camouflaged scene.

\noindent
\textbf{High-Frequency Detail Enhancer (HFDE).} 
To extract fine-grained edge and texture representations, we adopt the Laplacian pyramid decomposition, which is commonly used in related vision tasks \cite{lim2020dslr,liang2023low}. Specifically, the input image $I_0$ is decomposed as follows:

\begin{equation}
I_{k+1} = f^{k+1}_{\downarrow}(I_k), \quad 
{HF}_k = I_0 - f^{k+1}_{\uparrow}(I_{k+1}),
\end{equation}
where $k \in \{0, 1, 2, 3\}$ indicates the level of the Laplacian pyramid. The operators $f^{k}_{\downarrow}$ and $f^{k}_{\uparrow}$ denote downsampling and upsampling operations with a scale factor $2^k$. 

All residual components are then concatenated along the channel dimension to form the composite residual tensor 
${HF}_{\text{compose}} \in \mathbb{R}^{H \times W \times 12}$. Subsequently,  it is passed through the convolutional layer with a kernel size of $2 \times 2$ and a stride of 2, followed by layer normalization, to project it into a $C$-dimensional frequency embedding denoted as $HFE$.

To further enhance the multi-scale detail perception of high-frequency features, we stack six encoders hierarchically to construct a pyramid-like network that operates at progressively reduced resolutions, as illustrated in Figure \ref{framework} (b). Each encoder is formulated as:
\begin{equation}
y=\operatorname{ReLU}\left(\mathrm{ConvLN}\left(\operatorname{ReLU}\left(\mathrm{ConvLN}\left(x\right)\right)\right)+x\right),
\end{equation}
where $\mathrm{ConvLN}$ denotes a Convolutional layer with Layer Normalization. The lower-level encoders primarily preserve fine-grained edge and contour information, while the deeper layers capture more abstract and semantic structures.

Finally, we obtain a set of multi-scale and multi-level high-frequency representations 
$\{HFR_j^i \mid i,j=1,2,3; i \leq j\}$. 
Here, $i$ denotes the number of downsampling operations, and $j$ denotes the encoding stage. The shape of $HFR_j^i$ can be calculated as

\begin{equation}
HFR_j^i \in \mathbb{R}^{\frac{H }{2^{(i+1)}} \times \frac{W }{2^{(i+1)}}\times 2^i \cdot C}.
\end{equation}

\subsubsection{\textbf{Progressive Feature Fusion and Debiasing}}
To effectively bridge the cross-frequency representation gap and align multi-level high-frequency features at the same scale, we propose a Progressive Feature Fusion and Debiasing module. This module jointly exploits supervision signals from pseudo masks and scribble annotations, while adaptively mitigating label imbalance to achieve stable and unbiased optimization. The fused features are then forwarded to the hierarchical decoder and the scribble probability predictor, which collaboratively enable accurate camouflaged object segmentation and reliable label preference estimation.

\begin{algorithm}[]
\caption{Multi-scale and Multi-level Feature Fusion.}
\label{order}
\begin{algorithmic}[1]
\REQUIRE  Low- and high-frequency features $LFR^i, HFR_j^i$, frequency fusion function $\mathcal{F}$
\ENSURE  Fused features $\{F^i \mid i=1,2,3 \}$

\FOR{$i \gets 3 \text{ to } 1$}
    \STATE $L \gets LFR^i$
    \FOR{$j \gets 3 \text{ to } i$}
        \STATE $H \gets HFR_j^i$
        \STATE $L \gets \mathcal{F}(H, L)$
    \ENDFOR
    \STATE $F^i \gets L$
\ENDFOR
\end{algorithmic}
\end{algorithm}

\noindent
\textbf{Window-based Cross-Attention for Progressive Fusion.} 
Specifically, we first reshape and resize the low-frequency representation $LFR$ into a set of $\{LFR^i \mid i=1,2,3\}$ so that its dimensions match those of the high-frequency representations $\{HFR_j^i \mid i,j=1,2,3; i \leq j\}$. To gradually bridge the gaps among frequency representations of different spatial scales and semantic levels, we adopt a progressive fusion strategy, as summarized in Algorithm \ref{order}.

For the fusion function $\mathcal{F}(H, L)$, we employ a Window-based Cross-Attention that prioritizes local representation alignment over global interactions. For example, at the $i-th$ fusion level, the feature maps of $LFR^i$ and $HFR_j^i$ are divided into non-overlapping windows of size $w \times w$ to facilitate local semantic interaction. Inspired by the multi-head attention mechanism \cite{vaswani2017attention}, each window (e.g.,  $W_{lf}^k$ and $W_{hf}^k$) is flattened into a token sequence and projected through separate linear layers to obtain the query, key, and value vectors with dimension $d$: $\mathbf{Q}_{h},\mathbf{K}_{l},\mathbf{V}_{l}$. The attention weights within each window, representing the localized relationships between different frequency representations, are computed as:

\begin{equation}
    \mathbf{A}=\operatorname{Softmax}\left(\frac{\mathbf{Q}_{h} \mathbf{K}_{l}^{\top}}{\sqrt{d}}\right).
\end{equation}

Semantic cues from low-frequency values are aggregated as follows:

\begin{equation}
    \hat{W_{hf}^k}=\mathbf{A}\mathbf{V}_{l}.
\end{equation}

Finally, all $\hat{W_{hf}^k}$ are restored to their 2D structure, and the fused feature, which integrates local semantic guidance from low-frequency features into high-frequency details, is obtained via a skip connection and hierarchically fed into the CNN-based decoder for segmenting the camouflaged object at different scales.

\begin{figure}[]
    \centering
    \includegraphics[width=0.8\linewidth]{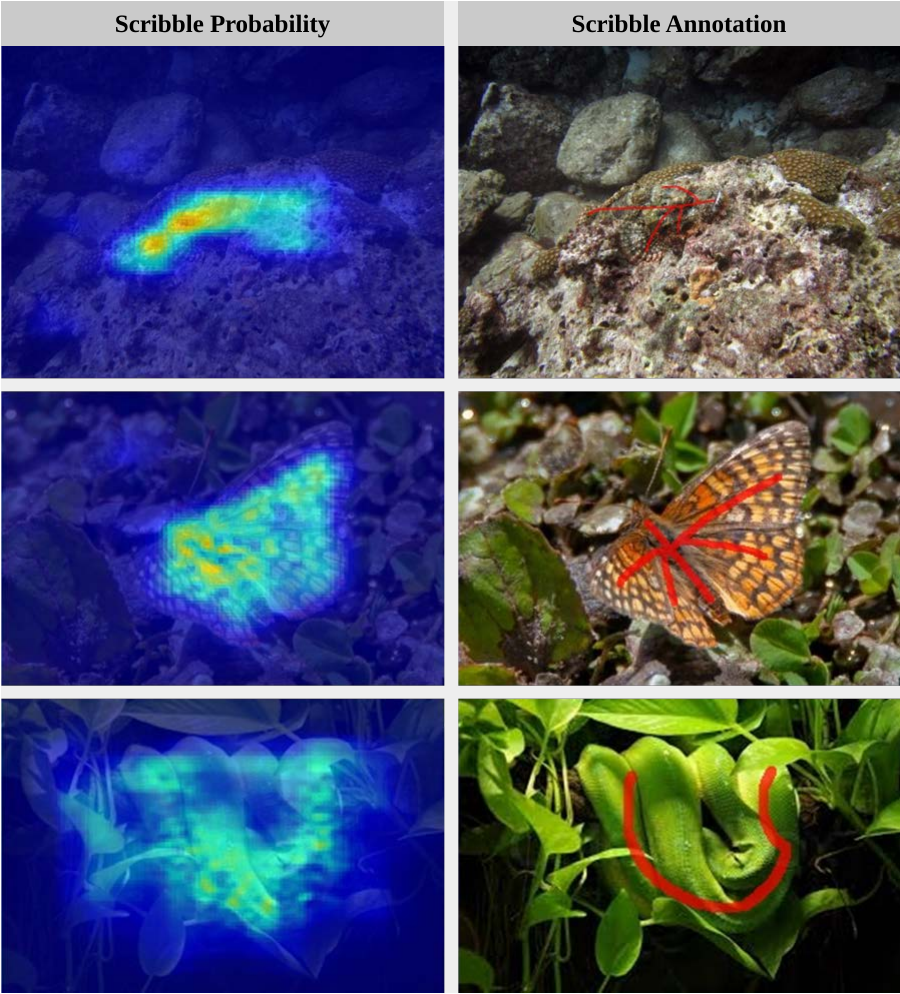}
    \caption{Visualization of the scribble probability map and the corresponding scribble annotation. The probability map highlights pixels that are most likely to be labeled by scribbles, although some of them may not be annotated in practice.}
    \label{vis-sp}
\end{figure}

\noindent
\textbf{Debiasing via Predicting Scribble Probability.} 
To mitigate the label imbalance inherent in scribble annotations, the first and foremost step is to estimate the probability distribution of scribble placements across each image. We adopt a two-layer ``Conv + BN + ReLU'' structure as the scribble probability predictor at different levels, whose prediction represents the probability of each pixel being labeled by scribbles or not, as visualized in Figure \ref{vis-sp}.

Here, we assume each image contains $N$ pixels, and the corresponding scribble label is denoted as $y \in \{0,1\}^{N}$, where a value of 1 indicates scribble regions and 0 denotes non-scribble regions. Accordingly, a weighted binary cross-entropy loss at the $k$-th ($k=1,2,3$) level is employed for pixel-level classification to compensate for the vast number of unlabeled pixels:

\begin{equation}
\mathcal{L}_{\text{Scrib}}^k = - \frac{1}{N} \sum_{i=1}^{N} \left[ \,\, y_i \log(p_{i,k}^{\text{scrib}}) + w_n \, (1 - y_i) \log(1 - p_{i,k}^{\text{scrib}}) \right],
\end{equation}
where $p_{i,k}^{\text{scrib}} \in [0,1]$ represents the estimated annotation probability of scribbles at pixel $i$, and $y_i \in \{0,1\}$ denotes the pixel-level ground-truth label for scribble classification. $w_n$ is the weight used to balance the contributions of the scribble and non-scribble classes during training.

Having established the predictor to estimate the scribble probability, the next step in alleviating harmful bias in scribbles is to dynamically adjust the supervision strength across regions during training. To this end, we introduce a debias loss derived from the Focal Loss \cite{lin2017focal}, which re-weights the segmentation supervision according to the predicted likelihood of each pixel being labeled in the scribble annotation. Specifically, for a segmentation prediction $p_{i,k}^{\text{seg}} \in [0,1]$ at pixel $i$ at the $k$-th ($k=1,2,3$) level, the Debias Loss is defined as:

\begin{equation}
\mathcal{L}_{\text{Debias}}^k = - \frac{1}{{N}^{\text{scrib}}} \sum_{i=1}^{N} y_i \big( 1 - p_{i,k}^{\text{scrib}} p_{i,k}^{\text{seg}} \big)^\gamma \log p_{i,k}^{\text{seg}},
\end{equation}
where ${N}^{\text{scrib}}$ denotes the number of pixels labeled by scribbles, and $\gamma$ controls the modulation strength.

Unlike the standard Focal Loss \cite{lin2017focal}, which employs $(1 - p_{i,k}^{\text{seg}})^{\gamma}$ as the modulation term, our formulation introduces a joint modulation factor $(1 - p_{i,k}^{\text{scrib}} p_{i,k}^{\text{seg}})^{\gamma}$ that adaptively reweights pixel-wise supervision based on both annotation reliability and prediction confidence. In this way, pixels with low scribble probability receive stronger supervision, while those with high annotation probability are down-weighted according to their segmentation confidence. 

By alleviating the harmful bias inherent in sparse and imbalanced scribble annotations, our $\mathcal{L}_{\text{Debias}}$ not only enables the model to better exploit the rich semantic cues contained in scribbles but also enhances its ability to capture the overall structure of camouflaged objects.

\noindent
\textbf{Objective Function.} 
Following previous works \cite{he2023camouflaged,jia2022segment}, we combine the commonly used binary cross entropy loss ($\mathcal{L}_\mathrm{BCE}$) and intersection-over-union loss ($\mathcal{L}_\mathrm{IoU}$). By progressively integrating multi-level supervision into the side outputs during the decoding process, the camouflaged object detection loss $\mathcal{L}_{\text{COD}}$ at the $k$-th ($k=1,2,3$) level can be expressed as:

\begin{equation}
\mathcal{L}_{\text{COD}}^k=\mathcal{L}_{\mathrm{BCE}}\left(P_{k}^{\text{seg}}, Y^{\text{mix}}\right)+\mathcal{L}_{\mathrm{IoU}}\left(P_{k}^{\text{seg}}, Y^{\text{mix}}\right), 
\end{equation}
where $P_{k}^{\text{seg}}$ denotes the segmentation output of the decoder at level $k$, and $ Y^{\text{mix}}$ represents the mixture of the pseudo masks obtained in Sec. \ref{s1} and the dilated scribble annotations.

Overall, the total loss function of our FADeNet is formulated as:

\begin{equation}
    \mathcal{L}_{\text{total}}=\sum_{k=1}^{3}(\mathcal{L}_{\text{COD}}^k+\alpha\mathcal{L}_{\text{Scrib}}^k+\beta \mathcal{L}_{\text{Debias}}^k),
\end{equation}
where $\alpha$ and $\beta$ are hyperparameters.

\section{Experiments}
\begin{table*}[]
\centering
\caption{Quantitative comparison with state-of-the-art methods on three benchmark datasets. “F” and “S” represent fully supervised and scribble-supervised labels, respectively. “–” denotes unavailable results. The best scores are highlighted in \textcolor{red}{red}, while the second-best ones are highlighted in \textcolor{blue}{blue}.}
\resizebox{\textwidth}{!}{
\begin{tabular}{l|l|c|cccc|cccc|cccc}
\toprule
\multirow{2}{*}{Methods} & \multirow{2}{*}{Publication} & \multirow{2}{*}{Label} &
\multicolumn{4}{c|}{CAMO} & \multicolumn{4}{c|}{COD10K} & \multicolumn{4}{c}{NC4K} \\

 & & & MAE $\downarrow$ & $S_m$ $\uparrow$ & $E_m$ $\uparrow$ & $F_{\beta}^{w}$ $\uparrow$
 & MAE $\downarrow$ & $S_m$ $\uparrow$ & $E_m$ $\uparrow$ & $F_{\beta}^{w}$ $\uparrow$
 & MAE $\downarrow$ & $S_m$ $\uparrow$ & $E_m$ $\uparrow$ & $F_{\beta}^{w}$ $\uparrow$ \\
\midrule
SINet~\cite{fan2020camouflaged} & CVPR 20 & F & 0.092 & 0.745 & 0.804 & 0.644 & 0.043 & 0.776 & 0.864 & 0.631 & 0.058 & 0.808 & 0.871 & 0.723 \\

UGTR~\cite{yang2021uncertainty} &ICCV 21 &F & 0.086 & 0.784 & 0.822 & 0.684 & 0.036 & 0.817 & 0.852 & 0.666 & 0.052 & 0.839 & 0.874 & 0.747 \\
MGL-R~\cite{zhai2021mutual} & CVPR 21&F & 0.088 & 0.775 & 0.812 & 0.673 & 0.035 & 0.814 & 0.851 & 0.666 & 0.052 & 0.833 & 0.867 & 0.740 \\
PFNet~\cite{mei2021camouflaged} & CVPR 21&F & 0.085 & 0.782 & 0.841 & 0.695 & 0.040 & 0.800 & 0.877 & 0.660 & 0.053 & 0.829 & 0.887 & 0.745 \\
UJSC~\cite{li2021uncertainty} &CVPR 21 &F & 0.073 & 0.800 &0.859& 0.728 &0.035& 0.809 &0.884& 0.684& 0.047& 0.842& 0.898& 0.771 \\

BSA-Net~\cite{zhu2022can} &AAAI 22 &F &0.079 & 0.794& 0.851&0.763&0.034& 0.818& 0.891& 0.738&0.048 & 0.841& 0.897&0.808\\
ZoomNet~\cite{pang2022zoom} & CVPR 22 &F & 0.066& 0.820& 0.892& 0.752& 0.029& 0.838& 0.911& 0.729& 0.043& 0.853& 0.896& 0.784 \\
SINetv2~\cite{fan2021concealed} &TPAMI 22 &F &0.070 & 0.820 &0.895 &0.782 &0.037 &0.815 & 0.906 &0.718 &0.048 & 0.847& 0.914&0.805\\

HitNet~\cite{hu2023high} &AAAI 23 &F &0.055& 0.849& 0.906& 0.831& \textcolor{blue}{0.023}&  0.871& 0.935& \textcolor{red}{0.823}& 0.037& 0.875& 0.926& 0.853 \\
SAM-Ada.~\cite{chen2023sam} &ICCV 23 &F & 0.070 &0.847& 0.873& 0.765& 0.025& 0.883& 0.918& 0.801& - & - & - & - \\
FEDER~\cite{he2023camouflaged} &CVPR 23 &F &0.071 &  0.802& 0.898& 0.781& 0.032& 0.822& 0.905& 0.751& 0.044& 0.808& 0.871& 0.769\\
FSPNet~\cite{huang2023feature}&CVPR 23 &F &0.050 &0.856& 0.899& 0.830& 0.026&  0.851& 0.895& 0.769& 0.035& 0.879& 0.915& 0.843\\

DINet~\cite{sun2024frequency} &ECCV 24  &F &0.068 &0.821 &0.874 &0.755 &0.031 &0.832 &0.903 &0.729 &0.043 &0.856 &0.909 &0.794\\
CamoDiffusion~\cite{chen2024camodiffusion} &AAAI 24 &F &\textcolor{blue}{0.042} &\textcolor{blue}{0.878} &\textcolor{red}{0.940} &\textcolor{blue}{0.853}  &\textcolor{red}{0.020} &\textcolor{red}{0.881} &\textcolor{red}{0.944} &\textcolor{blue}{0.814} &\textcolor{red}{0.029} &\textcolor{blue}{0.893} &\textcolor{blue}{0.942} &\textcolor{blue}{0.859}\\
CamoFormer~\cite{yin2024camoformer} &TPAMI 24 &F &0.046&  0.872& 0.929& \textcolor{red}{0.854}& \textcolor{blue}{0.023}& 0.869& 0.932& 0.811& \textcolor{blue}{0.030}&  0.891& 0.939& \textcolor{red}{0.868}\\
MCRNet~\cite{zhang2025mamba} &IJCV 25 &F &\textcolor{red}{0.041} &\textcolor{red}{0.883} &\textcolor{blue}{0.939} &0.849 &\textcolor{red}{0.020} &\textcolor{blue}{0.880} & \textcolor{blue}{0.941} & 0.812& \textcolor{red}{0.029} &\textcolor{red}{0.895} &\textcolor{red}{0.943} &0.857\\

\midrule

SCSOD~\cite{yu2021structure} &AAAI 21 &S & 0.102& 0.713& 0.795& 0.618& 0.055& 0.710& 0.805& 0.546 & - & - & - & - \\
CRNet~\cite{he2023weakly} & AAAI 23&S & 0.092 &0.735& 0.815 &0.641 &0.049& 0.733& 0.832& 0.576& 0.063& 0.775& 0.855& 0.688 \\
SAM~\cite{kirillov2023segment} &ICCV 23 & - & 0.132 &0.684& 0.687& 0.606& 0.050& 0.783& 0.798& 0.701& 0.078& 0.767 &0.776& 0.696 \\
SAM-S~\cite{kirillov2023segment} &ICCV 23 &S & 0.105 &0.731& 0.774& - &0.046& 0.772& 0.828& - &0.071& 0.763& 0.832& - \\
WS-SAM~\cite{hec2023weakly} &NeurIPS 23 &S & 0.092 &0.759 &0.818 &0.667 &0.038& 0.803& 0.878& 0.680 &0.052 &0.829& 0.886& 0.757 \\
SAM-COD~\cite{chen2024sam} &ECCV 24 &S &\textcolor{blue}{0.060} &\textcolor{blue}{0.836} &\textcolor{blue}{0.903} &\textcolor{blue}{0.779}& \textcolor{blue}{0.029} & \textcolor{blue}{0.833}& \textcolor{blue}{0.904} & \textcolor{blue}{0.728}& \textcolor{blue}{0.039}& \textcolor{blue}{0.859}& \textcolor{blue}{0.912}& \textcolor{blue}{0.795}\\
\rowcolor{mygray} \textbf{Ours}  & - &S & \textcolor{red}{0.045} &\textcolor{red}{0.868} &\textcolor{red}{0.928} & \textcolor{red}{0.823} & \textcolor{red}{0.024} & \textcolor{red}{0.863} & \textcolor{red}{0.929} & \textcolor{red}{0.775}  & \textcolor{red}{0.031}  & \textcolor{red}{0.890} & \textcolor{red}{0.938}& \textcolor{red}{0.842}\\

\bottomrule
\end{tabular}
}
\label{sota-cmp}
\end{table*}

\subsection{Experimental Setup}

\subsubsection{\textbf{Implementation Details}}
We implement our model using PyTorch \cite{paszke2019pytorch} and conduct all experiments on a single NVIDIA RTX 3090 GPU. In the first stage, we adopt SAM \cite{kirillov2023segment} with a ViT-H backbone and Qwen2.5-VL-7B \cite{Qwen2.5-VL} as the multimodal large language model (MLLM). In the second stage, following previous studies \cite{huang2023feature,xie2023frequency}, we employ the base version of ViT \cite{dosovitskiy2020image} pretrained with the DeiT strategy \cite{touvron2021training} as the transformer encoder.

For optimization, we use stochastic gradient descent (SGD) \cite{ruder2016overview} with a momentum of 0.9 and a weight decay of $5\times10^{-4}$. The batch size is set to 4, and the initial learning rate is 0.03. The learning rate is linearly warmed up during the first 20 epochs and subsequently decayed using a cosine annealing schedule. For loss configurations, the weight $w_n$ of the non-scribble class in $\mathcal{L}_{\text{Scrib}}$ is set to 0.02, and the modulation factor $\gamma$ in $\mathcal{L}_{\text{Debias}}$ is 0.9. In the overall loss formulation, the balancing coefficients are set to $\alpha = 2$ and $\beta = 0.5$.

\subsubsection{\textbf{Datasets}}
To evaluate the effectiveness of our proposed method, we leverage the scribble-annotated S-COD dataset \cite{he2023weakly}, which is specifically designed for weakly-supervised camouflaged object detection, to generate pseudo masks and train our model. The trained model is then evaluated on three widely used benchmark datasets: CAMO \cite{le2019anabranch}, COD10K \cite{fan2021concealed}, and NC4K \cite{lv2021simultaneously}. CAMO consists of 1,000 training images and 250 testing images. COD10K is a larger and more challenging dataset containing 3,040 training images and 2,026 testing images. NC4K serves as an additional benchmark, comprising 4,121 testing images without a training split.

\subsubsection{\textbf{Evaluation Metrics}}
Following existing works \cite{hec2023weakly,chen2024sam,he2023weakly}, we adopt four commonly used metrics for evaluation: mean absolute error (MAE), structure measure ($S_m$) \cite{fan2017structure}, mean E-measure ($E_m$) \cite{fan2021cognitive}, and weighted F-measure ($F_\beta^w$) \cite{margolin2014evaluate}.

\subsection{Comparison with State-of-the-art Methods}
\subsubsection{\textbf{Quantitative Comparison}}
To verify the effectiveness of our proposed framework, we compare it against state-of-the-art fully supervised and weakly-supervised methods, including SINet~\cite{fan2020camouflaged}, UGTR~\cite{yang2021uncertainty}, MGL-R~\cite{zhai2021mutual}, PFNet~\cite{mei2021camouflaged}, UJSC~\cite{li2021uncertainty}, BSA-Net~\cite{zhu2022can}, ZoomNet~\cite{pang2022zoom}, SINetv2~\cite{fan2021concealed}, HitNet~\cite{hu2023high}, SAM-Ada~\cite{chen2023sam}, FEDER~\cite{he2023camouflaged}, FSPNet~\cite{huang2023feature}, DINet~\cite{sun2024frequency}, CamoDiffusion~\cite{chen2024camodiffusion}, CamoFormer~\cite{yin2024camoformer}, MCRNet~\cite{zhang2025mamba}, SCSOD~\cite{yu2021structure}, CRNet~\cite{he2023weakly}, SAM~\cite{kirillov2023segment}, SAM-S~\cite{kirillov2023segment}, WS-SAM~\cite{hec2023weakly}, and SAM-COD~\cite{chen2024sam}.

From the results in Table \ref{sota-cmp}, it is evident that ${D}^{3}${ETOR} consistently outperforms all scribble-supervised baselines across all datasets. For example, on CAMO, it achieves an MAE of \textbf{0.045}, $S_m$ of \textbf{0.868}, $E_m$ of \textbf{0.928}, and $F_\beta^w$ of \textbf{0.823}, surpassing the second-best scribble-supervised method SAM-COD~\cite{chen2024sam} (MAE 0.060, $S_m$ 0.836, $E_m$ 0.903, $F_\beta^w$ 0.779) \textbf{by a clear margin.}

Furthermore, ${D}^{3}${ETOR} demonstrates competitive or superior performance compared to most fully supervised approaches, only slightly behind CamoDiffusion~\cite{chen2024camodiffusion} and MCRNet~\cite{zhang2025mamba} on certain metrics. Overall, these results indicate that ${D}^{3}${ETOR} establishes a new state-of-the-art for camouflaged object detection in weakly and even fully supervised scenarios using only scribble annotations.

\subsubsection{\textbf{Qualitative Comparison}}
\begin{figure*}[]
    \centering
    \includegraphics[width=\linewidth]{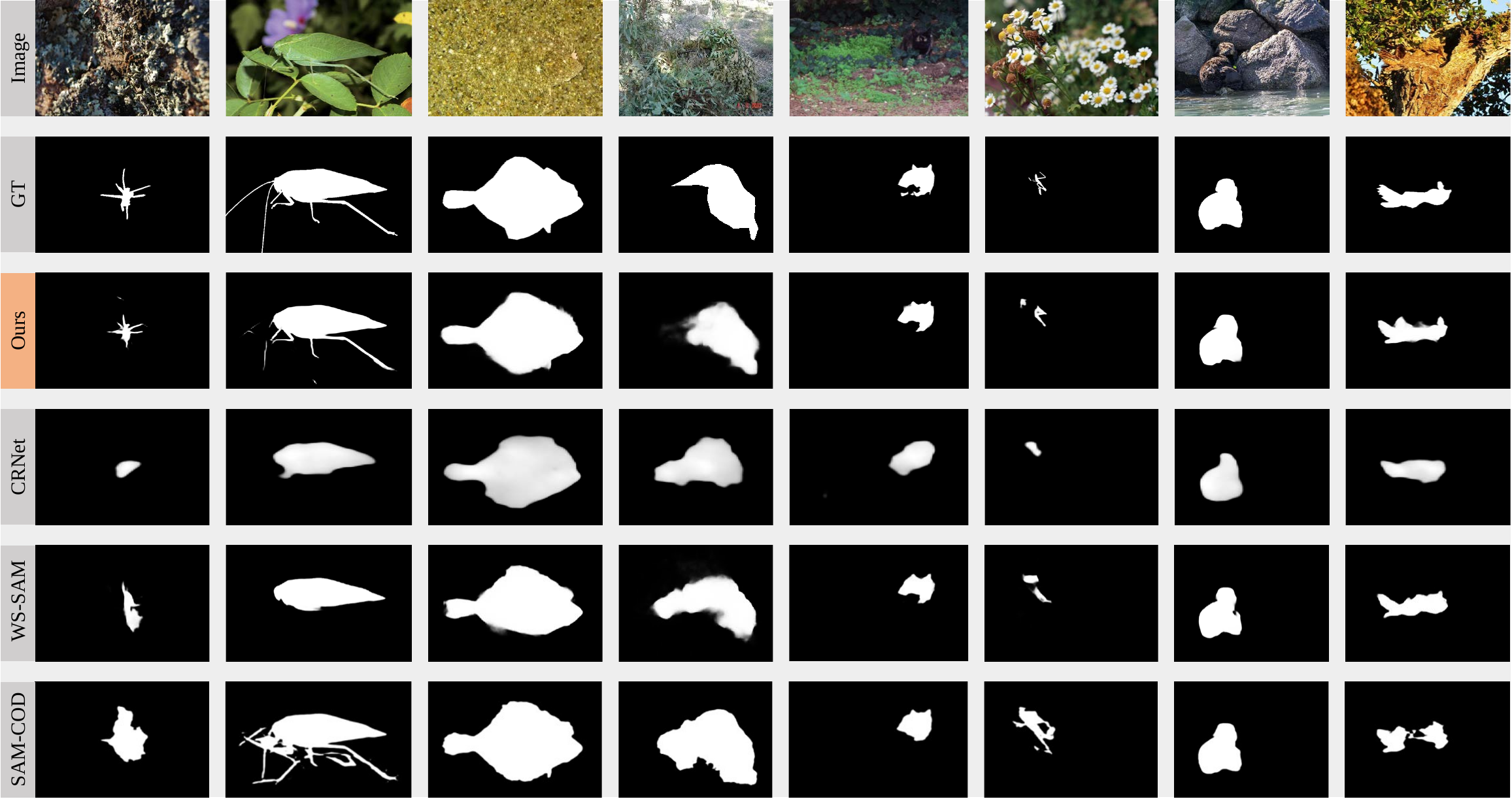}
    \caption{Qualitative comparison of our method with state-of-the-art scribble-based weakly-supervised methods under challenging scenarios.}
    \label{vis-cmp}
\end{figure*}

As illustrated in Figure~\ref{vis-cmp}, we select all existing weakly supervised approaches (i.e., CRNet~\cite{he2023weakly}, WS-SAM~\cite{hec2023weakly}, and SAM-COD~\cite{chen2024sam}) to visually compare against our proposed ${D}^{3}${ETOR}. The qualitative results indicate that ${D}^{3}${ETOR} accurately segments camouflaged objects under a variety of challenging conditions, including small-scale targets, intricate structural details, high visual resemblance to the background, fuzzy object boundaries, and complex surrounding contexts. In contrast, other methods fail to capture complete object regions or produce predictions with blurred details. These qualitative analyses provide an intuitive demonstration of the effectiveness and robustness of ${D}^{3}${ETOR}.

\subsection{Ablation Studies}

To verify the effectiveness of each main component and key designs within them, we perform comprehensive ablation studies on 8 different variants in our ${D}^{3}${ETOR}.

\subsubsection{\textbf{Impact of Debate-Enhanced Pseudo Labeling}}

As shown in Table~\ref{main-ab-s1}, to assess the contribution of each module in the pseudo labeling stage, we compare the generated pseudo masks with the ground truth in the S-COD training set and analyze several variants within the first stage.

\begin{table}[t]

\centering
\caption{Ablation analysis of the main components in Debate-Enhanced Pseudo Labeling on the training dataset S-COD.}

\begin{tabular}{cccccccc}
\hline
\multirow{2}{*}{} & \multirow{2}{*}{SAM} & \multirow{2}{*}{AEDPS} & \multirow{2}{*}{CoT} & \multicolumn{4}{c}{S-COD} \\
\cline{5-8}
 &  &  & & MAE $\downarrow$ & $S_m$ $\uparrow$ & $E_m$ $\uparrow$ & $F_{\beta}^{w}$ $\uparrow$ \\
\hline
\ding{172} & \checkmark &  & & 0.069 & 0.817 & 0.854 & 0.769 \\
\ding{173} & \checkmark & \checkmark &  & 0.037 & 0.858 & 0.909 & 0.829\\
ours & \checkmark &\checkmark  & \checkmark & 0.024 & 0.888 & 0.939 & 0.861 \\

\hline
\end{tabular}
\label{main-ab-s1}
\end{table}

\begin{itemize}
    \item In variant \ding{172}, only SAM is employed to generate pseudo masks, and all evaluation metrics indicate a notable deviation of its pseudo masks from the ground truth.
    
    \item Variant \ding{173} introduces the \textbf{Adaptive Entropy-Driven Point Sampling (AEDPS)} on top of variant \ding{172}, leading to substantial performance improvements across all metrics. Specifically, the MAE decreases from 0.069 to 0.037, while $F_\beta^w$ increases from 0.769 to 0.829. These results validate the effectiveness of AEDPS in prompting SAM to produce high-quality pseudo masks and mitigating the limited generalization capability of the SAM model in camouflaged object detection.
    
    \item When we further integrate the \textbf{Multi-Agent Debate with Multimodal Chain-of-Thought (CoT)} into variant \ding{173}, our method attains finer-grained semantic discrimination of pseudo masks and achieves the best results across all metrics. These results demonstrate the remarkable effectiveness of the debate-based reasoning strategy in filtering out noisy predictions and retaining more informative pseudo masks for subsequent training stages.
\end{itemize}

\subsubsection{\textbf{Impact of Frequency-Aware Progressive Debiasing}}
As illustrated in Table~\ref{main-ab-s2}, to quantify the contributions of the key components in the second stage, we conduct comprehensive ablation studies on CAMO, COD10K, and NC4K using several variants of our proposed \textbf{Frequency-Aware Debiasing Network (FADeNet)}.

\begin{table*}[]
\centering
\caption{Ablation analysis of the main components in Frequency-Aware Progressive Debiasing on the CAMO, COD10K and NC4K datasets}

\begin{tabular}{ccccccccccccc}
\hline
\multirow{2}{*}{} & \multirow{2}{*}{LFSE} & \multirow{2}{*}{HFDE}  & \multirow{2}{*}{WCAPF}  & \multirow{2}{*}{$\mathcal{L}_{\text{COD}}(mask)$} & \multirow{2}{*}{$\mathcal{L}_{\text{COD}}(mix)$} & \multirow{2}{*}{Debias} & \multicolumn{2}{c}{CAMO}
& \multicolumn{2}{c}{COD10K} & \multicolumn{2}{c}{NC4K} \\
\cline{8-13}
 &  &  &  & & &  & MAE $\downarrow$ & $S_m$ $\uparrow$  & MAE $\downarrow$ & $S_m$ $\uparrow$ &  MAE $\downarrow$ & $S_m$ $\uparrow$  \\
\hline
\ding{174} & \checkmark &  & & &\checkmark  &\checkmark  & 0.063 & 0.824  & 0.037 & 0.804  &0.051 &0.842 \\
\ding{175} & \checkmark & \checkmark & & &\checkmark &\checkmark  & 0.054 & 0.847  & 0.033 & 0.835 &0.047 &0.862 \\
\ding{176} & \checkmark &\checkmark  & \checkmark &\checkmark  & & & 0.051 & 0.850 &0.031 & 0.839 &0.042 &0.879 \\
\ding{177} & \checkmark &\checkmark  & \checkmark &  &\checkmark & & 0.049 & 0.854 &0.027 & 0.852 &0.041 &0.884 \\

ours & \checkmark &\checkmark  & \checkmark & &\checkmark &\checkmark & 0.045 &0.868 &0.024 & 0.863 &0.031 &0.890 \\
\hline
\end{tabular}
\label{main-ab-s2}
\end{table*}

\begin{figure}[]
    \centering
    \includegraphics[width=0.8\linewidth]{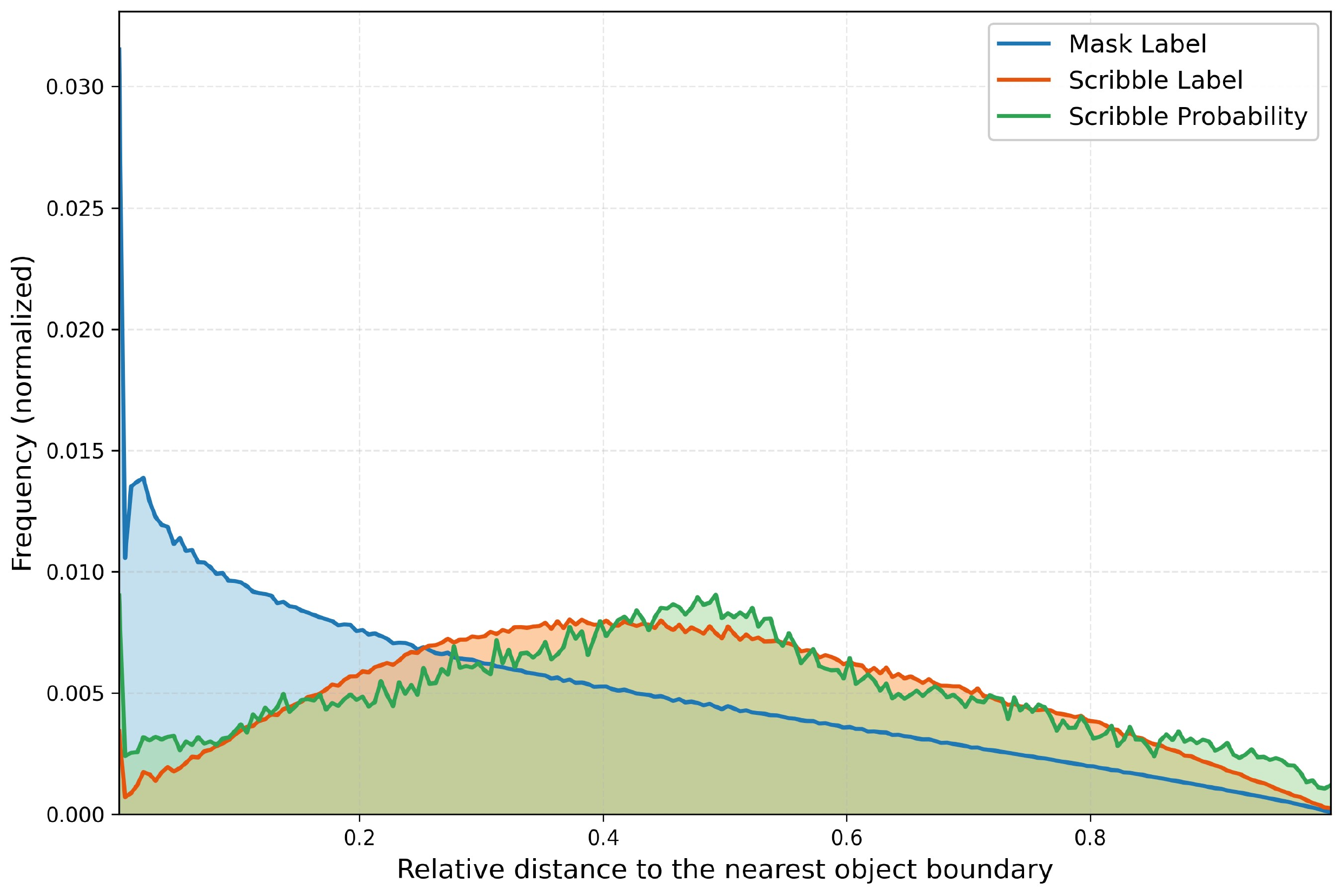}
    \caption{Distribution of relative distances from high-response pixels to their nearest object boundaries. The distribution of predicted scribble probability aligns closely with that shown in Figure \ref{dist}.}
    \label{dist-cmp}
\end{figure}

\begin{itemize}
    \item Variant \ding{174} includes only the \textbf{Low-Frequency Semantic Extractor (LFSE)}, which serves as the backbone of the detection network. The results reveal its limited capability in capturing fine details and maintaining structural consistency, indicating that low-frequency cues alone are insufficient for this task.
    
    \item By introducing the \textbf{High-Frequency Detail Enhancer (HFDE)}, variant \ding{175} achieves significant performance improvements across all datasets through the fusion of low- and high-frequency features via a simple cross-attention mechanism. The MAE decreases from 0.063 to 0.054 on CAMO, from 0.037 to 0.033 on COD10K, and from 0.051 to 0.047 on NC4K, while $S_m$ shows consistent gains (0.824→0.847 on CAMO, 0.804→0.835 on COD10K, 0.842→0.862 on NC4K). These results demonstrate that integrating high-frequency information effectively complements low-frequency semantics, enabling the model to delineate object boundaries more accurately and capture fine structural details essential for camouflaged object detection.

    \item After integrating the \textbf{Window-based Cross-Attention for Progressive Fusion (WCAPF)} into variant \ding{175}, the overall performance of variants \ding{176} and \ding{177} continues to improve. These findings indicate that WCAPF facilitates more effective multi-scale and multi-level frequency feature interactions, resulting in more precise segmentation of camouflaged objects. Furthermore, when the network is supervised solely by pseudo masks, as in variant \ding{176}, the performance gains in $S_m$ are limited compared with variant \ding{177}, which is jointly supervised by both pseudo masks and scribbles. This gap highlights the importance of incorporating scribble annotations for structural learning during training.

    \item Finally, we incorporate the \textbf{Debias} related losses to form FADeNet. Compared with variant \ding{177}, the complete network demonstrates consistent performance improvements, confirming that the debiasing process effectively mitigates label imbalance and enhances the discriminative capability of the frequency-aware representation.
    
     We also provide a qualitative comparison between the predicted scribble probability and the actual label preference. As shown in Figure \ref{dist-cmp}, the predicted probability distribution aligns closely with the scribble annotations, validating the effectiveness of our predictor in estimating scribble probabilities. Only when we have accurate scribble probability estimates can we adjust the supervision strength across different regions during training to mitigate inherent harmful bias.
\end{itemize}

\begin{table}[]
\centering
\caption{Ablation analysis of progressive feature fusion on the COD10K and NC4K datasets}

\begin{tabular}{cccccccc}
\hline
\multirow{2}{*}{} & \multirow{2}{*}{$F^1$} & \multirow{2}{*}{$F^2$}  & \multirow{2}{*}{$F^3$}  
& \multicolumn{2}{c}{COD10K} & \multicolumn{2}{c}{NC4K} \\
\cline{5-8}
 &  &  &    & MAE $\downarrow$ & $S_m$ $\uparrow$ & MAE $\downarrow$ & $S_m$ $\uparrow$ \\
\hline
\ding{174} &  &  &     & 0.037 & 0.804  &0.051 &0.842 \\
\ding{178} & \checkmark &  &     & 0.031 & 0.826  &0.039 &0.865 \\
\ding{179} & \checkmark & \checkmark &     & 0.025 & 0.857  &0.033 &0.886 \\
ours & \checkmark &\checkmark  & \checkmark   &0.024 & 0.863  &0.031 &0.890\\
\hline
\end{tabular}
\label{fusion}
\end{table}

\begin{figure}[]
    \centering
    \includegraphics[width=\linewidth]{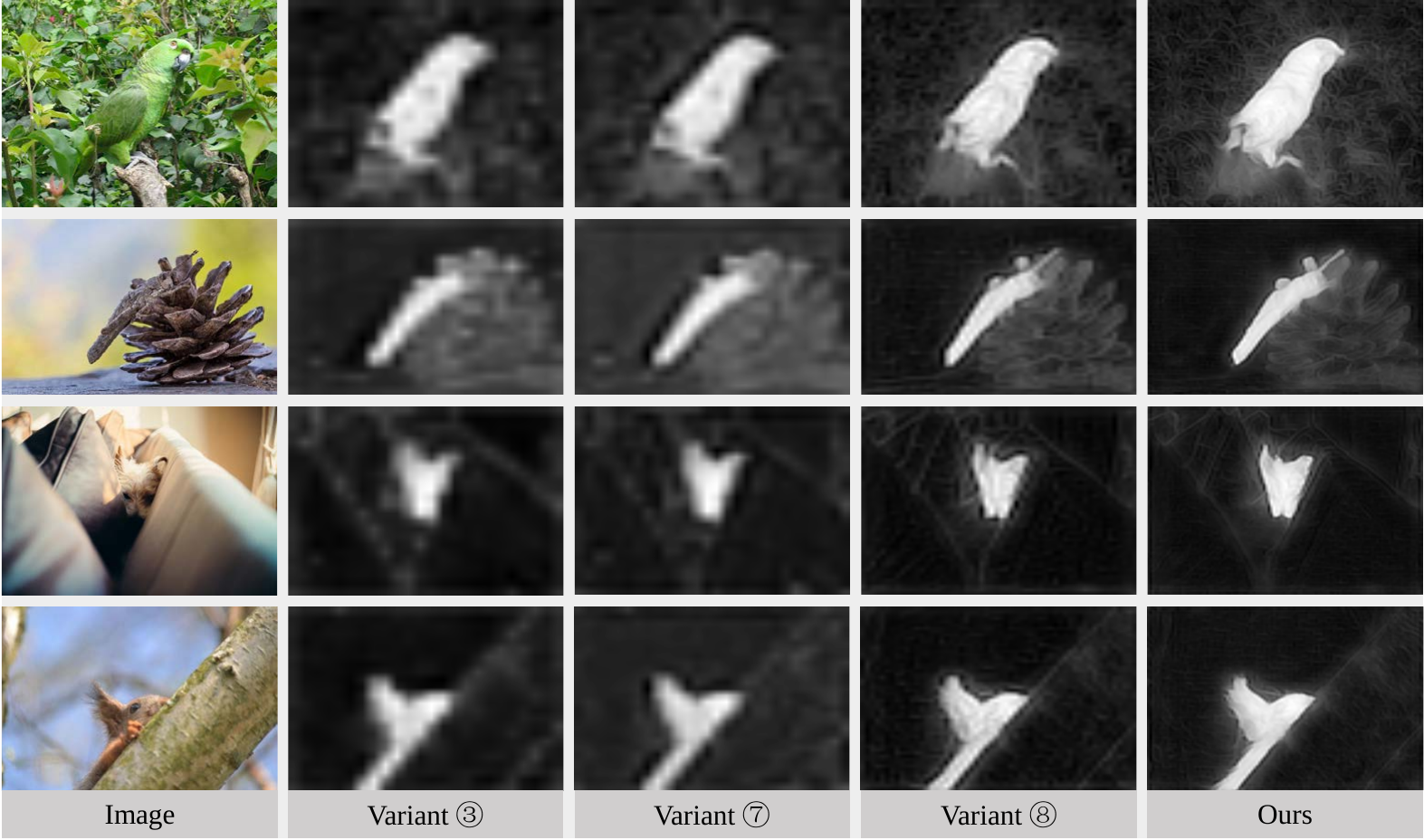}
    \caption{Visual comparison of feature maps obtained from different fusion levels within FADeNet, illustrating the progressive enhancement of detail and structure.}
    \label{vis-fusion}
\end{figure}

\subsubsection{\textbf{Impact of Progressive Feature Fusion}}
We further investigate the contributions of fused frequency-aware feature $F^i$ at different levels, as shown in Table~\ref{fusion}. The results indicate that integrating representations across multiple levels consistently enhances performance. This improvement arises from the complementary abstraction capabilities of features at different levels, which collectively strengthen the model’s ability to perceive camouflaged objects throughout the progressive fusion process.

Moreover, we visually compare the feature maps derived from $F^i$ at different fusion levels. As illustrated in Figure~\ref{vis-fusion}, these results clearly demonstrate that progressively fusing high-frequency information effectively captures object boundaries and refines fine structural details.

\section{Conclusion}

In this work, we presented ${D}^{3}$ETOR, a novel two-stage framework for Weakly-Supervised Camouflaged Object Detection with scribble annotations. By integrating Debate-Enhanced Pseudo Labeling and Frequency-Aware Progressive Debiasing, our method effectively addresses two fundamental challenges in WSCOD: the unreliability of pseudo masks generated by general-purpose models and the inherent bias in scribble annotations. Through adaptive entropy-driven point sampling and multi-agent debate, we enhance the capabilities of SAM to produce high-quality pseudo masks for COD. Building upon these pseudo masks and scribble annotations, we propose FADeNet to exploit complementary cues from low-frequency semantics and high-frequency structures, progressively fusing them via cross-attention while dynamically mitigating annotation bias through the debiasing loss. Extensive evaluations across multiple benchmark datasets demonstrate the effectiveness and robustness of ${D}^{3}$ETOR, substantially narrowing the gap with fully supervised methods.

\bibliographystyle{IEEEtran}
\bibliography{ref}

\end{document}